\documentclass[11pt,a4paper]{article}
\usepackage[hyperref]{acl2019}
\usepackage{times}
\usepackage{latexsym}

\usepackage{url}
\usepackage{multirow}
\usepackage{graphicx}
\usepackage{amsmath}
\usepackage{subfigure}
\usepackage{amsfonts}

\newcommand{\method}{{imitate}-NAT } 

\newcommand*{\affaddr}[1]{#1} % No op here. Customize it for different styles.
\newcommand*{\affmark}[1][*]{\textsuperscript{#1}}
\newcommand*{\email}[1]{\texttt{#1}}

\aclfinalcopy % Uncomment this line for the final submission
\def\aclpaperid{2166} %  Enter the acl Paper ID here

\title{Imitation Learning for Non-Autoregressive Neural Machine Translation}

\author{Bingzhen Wei\affmark[1]\thanks{This work was done when the first author was on an internship at Tencent.}, Mingxuan Wang, Hao Zhou, Junyang Lin\affmark[1,3], Jun Xie\affmark[4], Xu Sun\affmark[1,2] \\
\affaddr{\affmark[1]MOE Key Lab of Computational Linguistics, School of EECS, Peking University}\\
\affaddr{\affmark[2]Deep Learning Lab, Beijing Institute of Big Data Research, Peking University}\\
\affaddr{\affmark[3]School of Foreign Languages, Peking University}\\
\affaddr{\affmark[4]Tencent CSIG, BeiJing, China}\\
\email{\{weibz,linjunyang,xusun\}@pku.edu.cn} \\
\email{xuanxuans27@gmail.com, haozhou0806@gmail.com, stiffxie@tencent.com}\\
}

\date{}

\begin{document}
\maketitle
\begin{abstract}

Non-autoregressive translation models (NAT) have achieved impressive inference speedup.
A potential issue of the existing NAT algorithms, however, is that the decoding is conducted in parallel, without directly considering previous context.
In this paper, we propose an imitation learning framework for non-autoregressive machine translation, which still enjoys the fast translation speed but gives comparable translation performance compared to its auto-regressive counterpart. We conduct experiments on the IWSLT16, WMT14 and WMT16 datasets. Our proposed model achieves a significant speedup over the autoregressive models, while keeping the translation quality comparable to the autoregressive models.
By sampling sentence length in parallel at inference time, we achieve the performance of 31.85 BLEU on WMT16 Ro$\rightarrow$En and 30.68 BLEU on IWSLT16 En$\rightarrow$De.
\end{abstract}

\section{Introduction} \label{sec:intro}

Neural machine translation~(NMT) with encoder-decoder architectures~\citep{seq2seq, GRU} achieve significantly improved performance compared with traditional statistical methods\cite{phraseMT,statisticMT}.
Nevertheless, the autoregressive property of the NMT decoder has been a bottleneck of the translation speed. 
Specifically, the decoder, whether based on Recurrent Neural Network (RNN) \citep{LSTM, GRU} or attention mechanism \citep{transformer}, sequentially generates words. 
The latter words are conditioned on previous words in a sentence.
Such bottleneck disables parallel computation of decoder, which is serious for NMT, since the NMT decoding with a  large vocabulary is extremely time-consuming.

\begin{figure}[t!]
\centering
 
\subfigure[Autoregressive NMT]
{
	\begin{minipage}[b]{3.0cm}
	\centering         
	\includegraphics[scale=0.25]{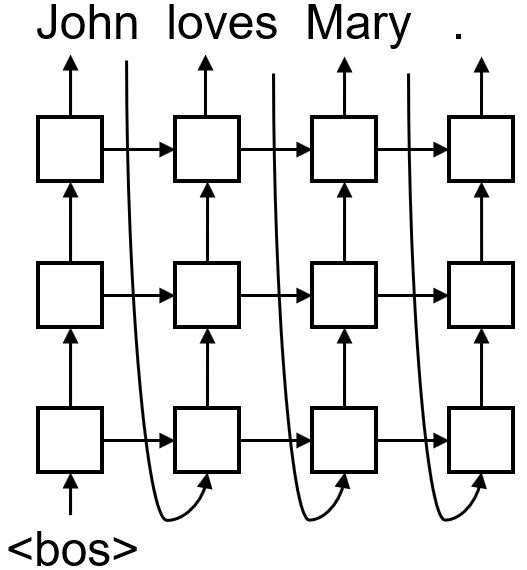}  
	\end{minipage}
}
\hfill
\subfigure[Non-Autoregressive NMT] 
{
	\begin{minipage}[b]{3.0cm}
	\centering  
	\includegraphics[scale=0.25]{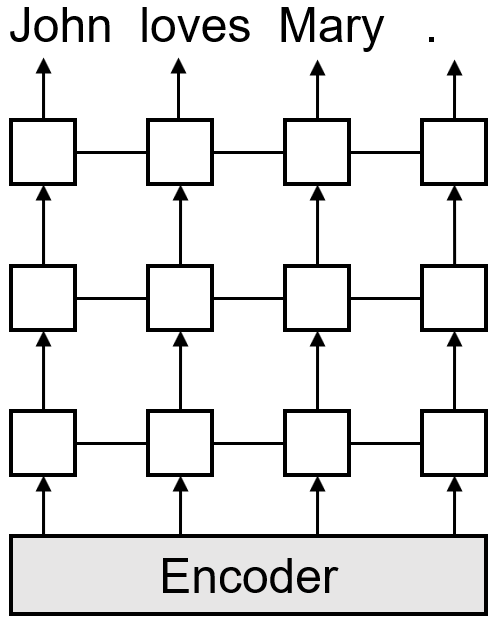} 
	\end{minipage}
}
 
\caption{ Neural  architectures for Autoregressive NMT and Non-Autoregressive NMT. }
\label{fig:intro}
\end{figure}

Recently, a line of research work~\citep{gu_non-autoregressive_2017, lee_deterministic_2018, libovicky_end--end_2018, wang_semi-autoregressive_2018} propose to break the autoregressive bottleneck by introducing non-autoregressive neural machine translation~(NAT).
In NAT, the decoder generates all words simultaneously instead of sequentially.
Intuitively, NAT abandon feeding previous predicted words into decoder state at the next time step, but directly copy source encoded representation~\cite{gu_non-autoregressive_2017, lee_deterministic_2018,guo_nat_decinp,2019Auxiliary} as inputs of the decoder. 
%This enables the decoder work in a parallel way, which significantly improves the decoding speed of NMT with moderate accuracy loss~(always within 5 BLEU).
Thus, the generation of the NAT models does not condition on previous prediction.
NAT enables parallel computation of decoder, giving significantly fast translation speed with moderate accuracy 
~(always within 5 BLEU). Figure~\ref{fig:intro} shows the difference between autoregressive and non-autoregressive models.
% \TODO{we can even add more related work here.} 
%%%%%%%%%%%%%%%%%%%%%%%%%%%%%%%%%%%%%%%%%
%%%%%%%%%%%%%%%%%%%%%%%%%%%%%%%%%%%%%%%%%
%\begin{figure}[t]
%\centering
%\footnotesize
%\includegraphics[width=0.9\linewidth]{fig/NAT_illustration.png}
%\caption{More efforts are needed to make this illustration more clear. The %point is NAT does not feed predicted word into decoder at the next step.}
%\label{fig:intro}
%
%\end{figure}

However, we argue that current NAT approaches suffer from delayed supervisions~(or rewards) and large search space in training.
NAT decoder simultaneously generates all words of the translation, the search space of which is very large.
For one time step, decoding states across layers~(more than 16 layers) and time steps could be regarded as a 2-dimensional sequential decision process.
Every decoding state has not only to decide which part of target sentence it will focus on, but also to decide the correct target word of that part.
All decisions are made by interactions with other decoding states.
Delayed supervisions~(correct target word) will be obtained by decoding states in the last layer, and intermediate decoding states will be updated by gradient propagation from the last layer.
Therefore, the training of NAT is non-trivial and it may be hard for NAT to achieve a good model, which is the same case that reinforcement learning~\cite{Mnih2013PlayingAW,Mnih2015HumanlevelCT} is hard to learn with large search space.
The delayed supervision problem is not severe for autoregressive neural machine translation(AT) because it predicts words sequentially.
Given the previous words, contents to be predicted at current step are relatively definite, thus the search space of AT is exponentially lower than NAT.
We blame the delayed supervision and large search space for the performance gap between NAT and AT.

In this paper, we propose a novel imitation learning framework for non-autoregressive NMT~(\method ).
Imitation learning has been widely used to alleviate the problem of huge search space with delayed supervision in RL.
It is straightforward to bring the imitation learning idea for boosting the performance of NAT.
Specifically, we introduce a knowledgeable AT demonstrator to supervise each decoding state of NAT model.
In such case, 
Specifically, We propose to employ a knowledgeable AT demonstrator to supervise every decoding state of NAT across different time steps and layers, which works pretty well practically.
Since the AT demonstrator is only used in training, our proposed \method enjoys the high speed of NAT without suffering from its relatively lower translation performance. 

Experiments show that our proposed \method is fast and accurate, which effectively closes the performance gap between AT and NAT on several standard benchmarks, while maintains the speed advantages of NAT~(10 times faster).
On all the benchmark datasets, our \method with LPD achieves the best translation performance, which is even  close to the results of the autoregressive model.

\section{Background}
In the following sections, we introduce the background about Autoregressive Neural Machine Translation and Non-Autoregressive Neural Machine Translation.

\subsection{ Autoregressive Neural Machine Translation}
Sequence modeling in machine translation has largely
focused on autoregressive modeling which generate
a target sentence word by word from left to
right,
denoted by $p_\theta(Y|X)$, where  
$X=\{x_1\cdots,x_T\}$ and
$Y=\{y_1,\cdots,y_{T'} \}$
represent the source and target sentences
as sequences of words respectively.
$\theta$ is
a set of parameters  usually trained to minimize
the  negative loglikelihood:%  inconditional likelihood of the correct translation $Y$ given the source sentence $X$,i.e.,:
\begin{eqnarray}
      %\mathcal{L}_{AT}= \prod_{i=1}^{T'} p(y_i|y_{<i},X).
      \mathcal{L}_{AT}= -\sum_{i=1}^{T'} \log p(y_i|y_{<i},X).
      \label{eq:AT}
\end{eqnarray}

where $T$ and $T'$ is the length of the 
source and the target sequence respectively.

Deep neural network with 
autoregressive framework has
achieved great success on machine translation, 
with different choices of architectures.
The RNN-based NMT approach, or RNMT, was
quickly established as the de-facto standard for
NMT. Despite the recent success, the inherently sequential architecture
prevents RNMTs from being parallelized during training and inference.
Following RNMT, CNNs
and self-attention based models have recently drawn research attention
due to their ability to fully parallelize training to
take advantage of modern fast computing devices.
However, the autoregressive nature
still creates a bottleneck at inference stage, since without
ground truth, the prediction of each target token has to condition on previously predicted tokens. 
%%%%%%%%%%%%%%% Model Figure %%%%%%%%%%%%%%%%%
\begin{figure*}[t]
\centering
\footnotesize
\includegraphics[width=0.8\linewidth]{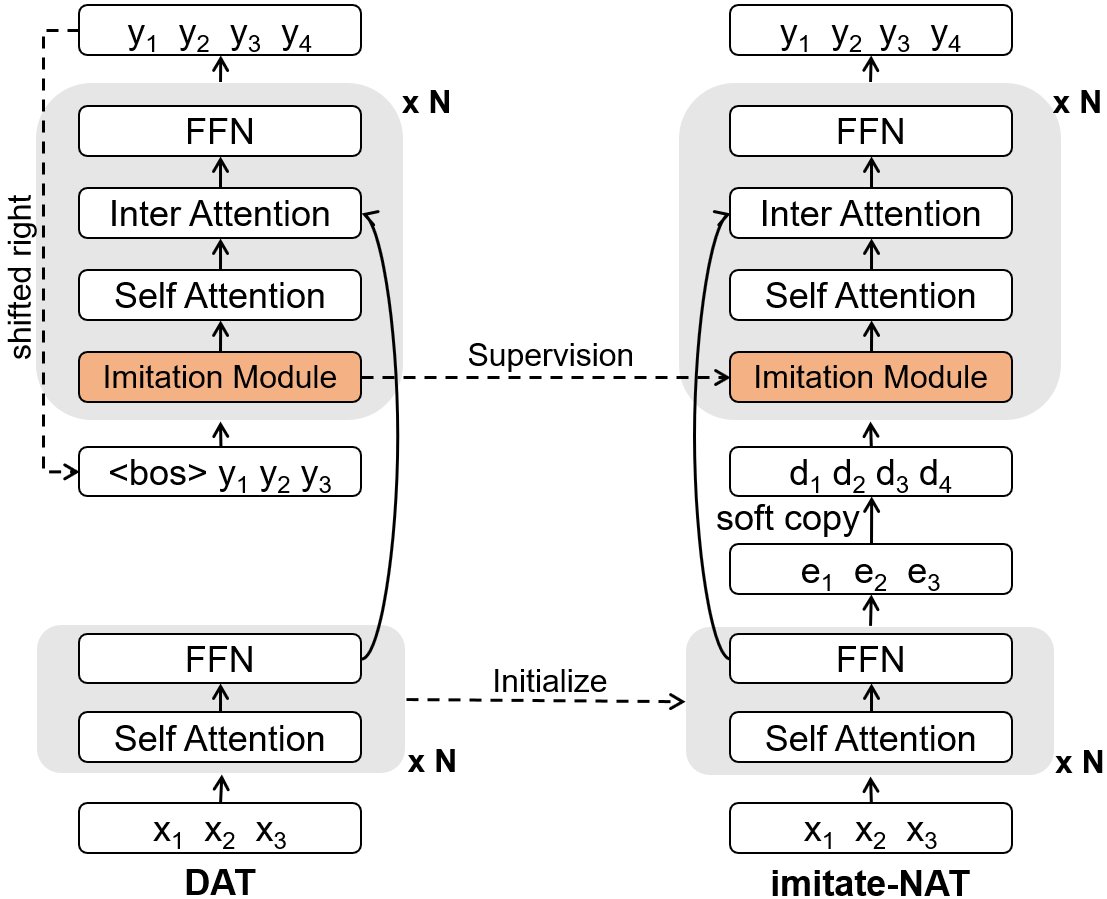} %model-crop.pdf
\caption{Illustration of the proposed model, where the black solid arrows represent differentiable connections and the dashed arrows are non-differentiable operations. Without loss of generality, this figure shows the case of T=3, T'=4. The left side of the figure is the DAT model and the right side is the \method. The bottom is the encoder and the top is the decoder. The internal details of Imitation Module are shown in Figure~\ref{fig:action}.}%Since the upper layer has no dependency on UCL, AT-C model is equivalent to the transformer model during the inference.}
\label{fig:model}
\end{figure*}
%%%%%%%%%%%%%%%%%%%%%%%%%%%%%%%%%%%%%%%%%%%%%%

\subsection{Non-Autoregressive Neural Machine Translation}
As a solution to the issue of slow decoding,
~\citet{gu_non-autoregressive_2017}  recently proposed non-autoregressive model (NAT) 
 to break the inference bottleneck by exposing all decoder inputs to the network simultaneously.
 NAT removes the autoregressive connection directly and factorizes the
target distribution
into a product of conditionally independent per-step distributions.   The negative loglikelihood loss function for NAT model become is then defined as:
\begin{equation}
   %\mathcal{L}_{NAT}= p(T'|X)\prod_{i=1}^{T'} p(y_i|X).
   \mathcal{L}_{NAT}= -\sum_{i=1}^{T'} \log p(y_i|X).
   \label{eq:LNAT}
\end{equation}
The  approach breaks the dependency among the target words across time,
thus the target distributions can be computed
in parallel at inference time.

In particular,  the encoder stays unchanged from the original Transformer network. A latent fertility model is then
used to
copy the sequence of source embeddings as the input of the decoder.
The decoder has the same architecture as the encoder plus the encoder attention.
The best results were achieved by sampling fertilities from
the model and then rescoring the output sentences
using an autoregressive model. 
The reported inference speed of this method is 2-15 times faster than a comparable autoregressive model, depending
on the number of fertility samples.

This desirable property of exact and parallel decoding however comes at the expense of potential performance degradation.
Since the conditional
dependencies within the target sentence ($y_t$ depends on
$y_{<t}$) are removed from the decoder input, the decoder is
not powerful enough to leverage the inherent
sentence structure for prediction.
Hence the decoder has to figure out such target-side information by itself
just with the source-side information
during training, which leads to a larger modeling gap between the  true model and the
neural sequence model. 
Therefore, strong supervised signals could be introduced as the latent variable to help the model learn better
internal dependencies within a sentence.

In AT models, the generation of the current token 
is conditioned on previously generated tokens
, which provides strong target side context
information. In contrast, NAT models generate tokens
in parallel, thus the target-side dependency  is indirect and weak.
Consequently, the decoder of a NAT model has to handle the
translation task conditioned on less and weaker information
compared with its AT counterpart, thus leading to inferior
accuracy.

\section{Proposed Method: \method}

In this section, we propose an imitation learning framework~(\method) to close the performance gap between the NAT and AT.

%%%%%%%%%%%%%%%%%%%%%%%%%%%%%%%%%%%%%%%%%
\begin{figure}[t!]
\centering
 
\subfigure[Imitation module of DAT]
{
	\begin{minipage}[b]{3cm}
	\centering         
	\includegraphics[scale=0.2]{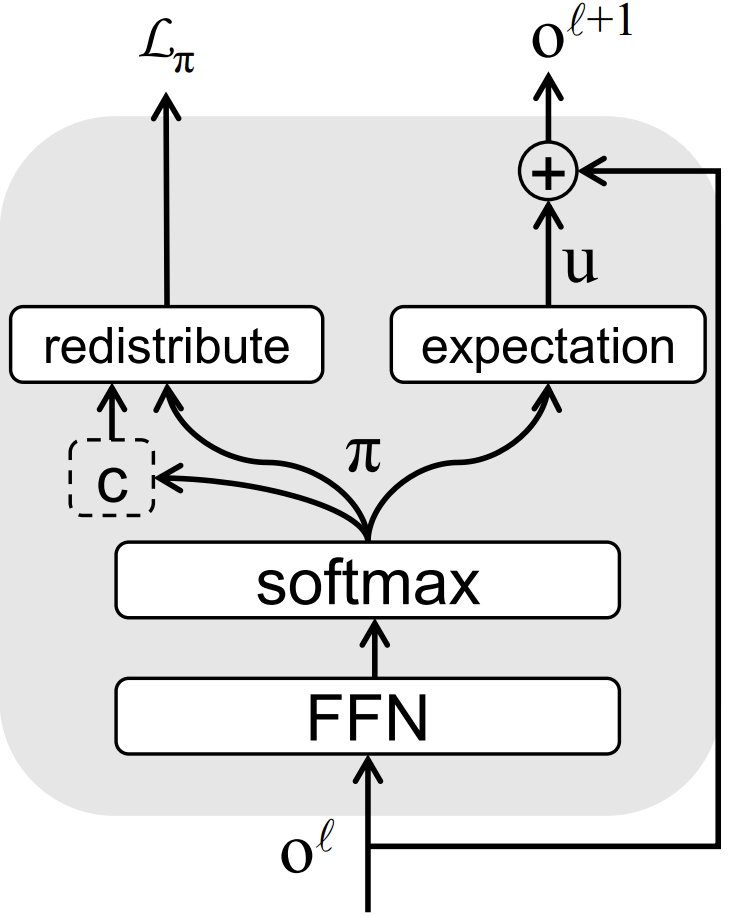}  %at_cropped.pdf
	\end{minipage}
}
\hfill
\subfigure[Imitation module of \method] 
{
	\begin{minipage}[b]{3cm}
	\centering  
	\includegraphics[scale=0.2]{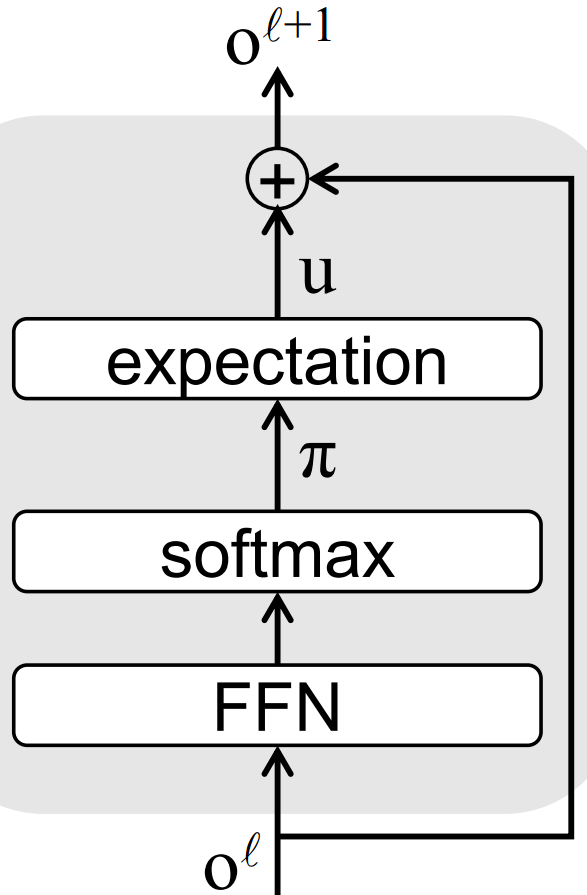} %nat_cropped.pdf 
	\end{minipage}
}
 
\caption{The imitation module of AT demonstrator and NAT learner.}
\label{fig:action}
\end{figure}
%%%%%%%%%%%%%%%%%%%%%%%%%%%%%%%%%%%%%%%%%

\subsection{Preliminary of \method}

We bring the intuition of imitation learning to non-autoregressive NMT and adapt it to our scenario.
Specifically, the NAT model can be regarded as a \textit{learner}, which will imitate a knowledgeable \textit{demonstrator} at each decoding state across layers and time steps.
However, obtaining an adequate demonstrator is non-trivial.
We propose to employ an autoregressive NMT model as the demonstrator, which is expected to offer efficient supervision to each decoding state of the NAT model.
Fortunately, the AT demonstrator is only used in training, which guarantees that our proposed \method enjoys the high speed of NAT model without suffering from its relatively lower performance.

In following parts, we will describe the \textit{AT demonstrator} and the \textit{NAT learner} in our \method framework, respectively.

% supervised by a demostrator 

% A classic family of imitation learning methods is to collect data from running the demonstrator and train a regressor or
% classifier via supervised learning.
% These methods  learn
% a policy $\hat{\pi}^*$ from a fixed-size dataset precollected from the demonstrator or expert.
% In this paper, we train the NAT model  in the imitation learning (IL) framework
% surprivised by the relevant action-level signals from an AT demonstrator.

\subsection{AT Demonstrator}
\label{sec:AT}
For the proposed AT, we apply a variant of the transformer model as the demonstrator, named DAT.
The encoder stays unchanged from the original Transformer network.
A crucial difference 
lies in that the decoder introduces the imitation module which emits actions
at every time step.
The action brings sequential information,  thus can be used as the guidance signal during the NAT
training process.

The input of each decoder layer $O^\ell=\{o_1^\ell,o_2^\ell,
\cdots,o_{T'}^\ell\}$
 can be
considered as the observation (or environment) of
the IL framework, where $\ell$ donates the layer of the observation.
Let $A^\ell=\{a_1^\ell,a_2^\ell,\cdots,a_{T'}^\ell\} \in \mathcal{A}$ denotes an action sequence from the action space $\mathcal{A}$. The action space $\mathcal{A}$ is finite and its size $n$ is a hyperparameter, representing the number of action categories.
The distribution of the action of DAT can be then fed to the NAT model as the training signal.
Let $\Pi$ denotes  a
policy class, where each $\pi^\ell \in \Pi$ generates an action distribution sequence $A^\ell$ in response to a context sequence
$O^{\ell}$.

Predicting actions $A^\ell$ may depend on the contexts of previous layer
$O^{\ell}$ and policies $\pi^\ell$ can thus be viewed as mapping states to actions.
 A roll-out of $\pi$ given the context sequence $O^{\ell}$
 to determine the action sequence $A^{\ell}$, which is:
 \begin{equation}
      a_t=\arg\max(\pi^\ell(o_t^{\ell}))
 \end{equation}
 where
 \begin{equation}
     \pi^\ell(o_t^{\ell}) = \text{softmax}(\text{FFN}(o_t^{\ell})).
 \end{equation}
 The distribution  $ \pi^\ell(o_t^{\ell})$ represents the probability
 of the decision depends on the current state or environment $o_t^{\ell}$.
 The discrete operation $\arg\max(\cdot)$ suffers from
 the non-differentiable problem which makes it
 impossible to train the policy from an end to end framework.
 
Note that unlike the general reinforcement or
imitation learning framework, 
we consider to compute the action state which as the
expectation of  the embedding of the action $a_t$:
\begin{eqnarray}
  u^\ell_t =  \mathbb{E}_{a_t^\ell\sim  \pi^\ell(o_t^{\ell})} \delta(a_t^\ell),
\end{eqnarray}
where $\delta(a_t^\ell)\in \mathbb{R}^k$ returns the  embedding of the action $a_t^\ell$ and $k$ denotes the embedding dimension.
The states of next layer are then based on the current output of the  decoder state and the emitted action state:
\begin{equation}
    %o_t^{\ell+1} =   u^\ell_t + \text{Transfer}(o_t^{\ell}),
    o_t^{\ell+1} =  \text{Transfer}(u^\ell_t + o_t^{\ell}),
\end{equation}
where $\text{Transfer}(\cdot)$ denotes the 
vanilla
transformer decoding function including a self-attention layer, an encoder-decoder attention layer and followed by a FFN layer~\cite{transformer}.

\subsubsection{Action Distribution Regularization}
\label{sec:ADR}
The supervised signal for the action distribution $\pi(o_t)$
is not direct in NAT, 
thus the action prediction can be viewed as an  unsupervised clustering problem.
One potential issue is the unbalanced distribution of action.
Inspired by \citet{Xie2016UnsupervisedDE}, we  introduce a regularization method to increase the space utilization.
Formally, an moving average $c$ is applied to calculate the cumulative activation level for each action category:
\begin{equation} \label{eq:re1}
    c \leftarrow \alpha\cdot c + (1-\alpha)\sum^{T'}_{t=1} \pi(o_t)/T'
\end{equation}
We set $\alpha$ 0.9 in our experiments.
Then $\pi'(o_i)$ can be re-normalized with 
the cumulative history $c$:
\begin{equation} \label{eq:re2}
\pi'(o_t) = \frac{\pi(o_t)^2/ c}{\sum_{j}\pi(o_t)_j^2/ c_j}
\end{equation}
The convex property of the quadratic function can adjust the distribution to achieve the purpose of clustering. The role of $c$ is to redistribute the probability distribution of $\pi(o_t)$, which leads to a more balanced category assignment. 

We define our objective as a KL divergence loss between $\pi(o_t)$ and the auxiliary distribution $\pi'(o_t)$ as follows:
\begin{equation}
\mathcal{L}_{\pi} = \sum_t \pi'(o_t)\log \frac{\pi'(o_t)}{\pi(o_t)}
\end{equation}

\subsection{NAT learner}

\subsubsection{Soft Copy}
To facility the imitation learning process, our \method 
is based on the 
AT demonstrator described in section~\ref{sec:AT}.
The only difference lies in that 
the initialization of the decoding inputs.
Previous approaches apply a UniformCopy method to address the problem. More specifically, the decoder input at position $t$ is the copy of the encoder 
embedding at position  $Round(T't/T)$ ~\cite{gu_non-autoregressive_2017,lee_deterministic_2018}. 
As the source and target sentences are often of different lengths, AT model  need 
to predict the target length $T'$ during inference stage.
The length prediction problem can be viewed as a typical
classification problem 
based on the output of the encoder.
we follow \citet{lee_deterministic_2018} 
to predict the length of the target 
sequence.

The proposed $Round$ function is unstable and non-differentiable, which make the decoding task difficult. 
We therefore propose a  differentiable and robust method named SoftCopy 
 following the spirit of the attention mechanism~\cite{humanreading,bengio}. 
 The weight $w_{i,j}$  depends on the distance relationship between the source position $i$  and the target position $j$. 
 \begin{equation}
      w_{ij} = \text{softmax}(-|j-i|/\tau)
 \end{equation}
$\tau$ is a trainable parameters used to adjust the degree of focus when copying.
Then the input of the target at position 
$j$ can be computed as :
\begin{equation}
    y_j = \sum_{i=0}^{T} w_{ij}x_i,
\end{equation}
where $x_i$ is usually the source embedding at position $i$. 
It is also worth mentioning that we take
the top-most hidden states instead of the 
word embedding as $x_i$ in order to cache the global context information. 

\subsubsection{Learning from AT Experts }
The conditional independence assumption
prevents NAT model from
properly capturing the highly multimodal 
distribution of target translations. 
AT models takes already generated target
tokens as inputs, thus can provide complementary extension information for 
NAT models.
A straightforward idea to bridge the gap 
between NAT and AT
is that NAT can actively learn the behavior of AT step by step.

The AT demonstrator generate
action distribution
$\pi_{AT}(O) \in \mathbb{R}^n$ 
as the posterior supervisor signal. 
We expect the supervision information can guide the generation process of NAT.
The \method exactly follows the same decoder structure with our AT demonstrator, and  
emits distribution $\pi_{NAT}(O) \in \mathbb{R}^n$  to learn from AT demonstrator step by step.
More specifically, we try to minimize 
the cross entropy of the distributions between the two policies:
%\begin{eqnarray}
%    \mathcal{L}_{IL}&=& H(\pi_{NAT}(o_t),\pi_{AT}(o_t)) \\
%    &=& \mathbb{E}_{\pi_{NAT}(o_t)}\log \pi_{AT}(o_t)
%\end{eqnarray} \label{loss_post}
\begin{eqnarray}
    \mathcal{L}_{IL}&=& H(\pi_{AT}(o_t),\pi_{NAT}(o_t)) \\
    &=& -\mathbb{E}_{\pi_{AT}(o_t)}\log \pi_{NAT}(o_t)
\end{eqnarray} \label{loss_post}

\begin{table*}[htb]
\centering

	\begin{tabular}{l| c c c c c c}
		\hline
		 \multirow{2}{*}{Models} &  \multicolumn{2}{c}{WMT14}              & \multicolumn{2}{c}{WMT16}              & IWSLT16  \\ 
		                         & En$\rightarrow$De & De$\rightarrow$En   & En$\rightarrow$Ro  &Ro$\rightarrow$En   &En$\rightarrow$De & Speedup \\
		 \hline
		   Transformer~\cite{transformer} & 27.41 &   31.29                   & / & /                    &30.90 &1.00$\times$ \\
		                      AT Demonstrator & 27.80 &   31.25                   & 33.70 & 32.59                     &30.85 &1.05$\times$ \\
        \hline
        NAT-FT\citep{gu_non-autoregressive_2017} & 17.69 & 21.47  &27.29 &29.06 &26.52 &15.60$\times$ \\
        NAT-FT(+NPD s=10)    &18.66 &22.41  &29.02 &30.76 &27.44 &7.68$\times$ \\
        NAT-FT(+NPD s=100)   &19.17 &23.20  &29.79 &31.44 &28.16 &2.36$\times$ \\
        NAT-IR($i_{dec}=1$)  &13.91 &16.77  &24.45 &25.73 &22.20 &8.90$\times$ \\
        NAT-IR($i_{dec}=10$) &21.61 &25.48  &29.32 &30.19 &27.11 &1.50$\times$ \\
        LT                   &19.80 &/      &/     &/     &/     &5.78$\times$ \\
        LT(rescoring 10)     &21.0  &/      &/     &/     &/     &/ \\
        LT(rescoring 100)    &22.5  &/      &/     &/     &/     &/ \\
        \hline
        \textbf{NAT without imitation} & 19.69    & 22.71    &     /       &   /    &25.34      &18.6$\times$          \\
       \textbf{\method}       &   22.44  & 25.67    &28.61       &28.90  &28.41 &18.6$\times$  \\
        \textbf{\method}(+LPD,$\Delta T=3$)   &  24.15   &  27.28   &31.45   &31.81   & 30.68 &9.70$\times$  \\
        %\textbf{\method}(+SPD+LPD) &\textbf{} &\textbf{}  &\textbf{31.74/5.3$\times$}  &\textbf{32.11/5.5$\times$}  &\textbf{29.76/4.99$\times$} \\
        \hline
	\end{tabular}
	\caption{The test set performances of AT and NAT models in BLEU score. NAT-FT, NAT-IR and LT  denotes the competitor method
in ~\cite{gu_non-autoregressive_2017}, ~\cite{lee_deterministic_2018} and   ~\cite{kaiser_fast_2018} respectively. \method is our proposed NAT with imitation learning.
	\label{tab:bleu}}

\end{table*}
%%%%%%%%%%%%%%%%%%%%%%%%%%%%%%%%%%
\subsection{Training}
In the training process, the action distribution regularization term  described in 
~\ref{sec:ADR} is
combined with the commonly used cross-entropy loss in ~Eq.~\ref{eq:AT}:
\begin{equation}
    \mathcal{L}_{AT}^* = \mathcal{L}_{AT} + \lambda_1 \mathcal{L}_{\pi}
\end{equation}

For NAT models,  the imitation learning  term are combined with the 
 commonly used cross-entropy loss in ~Eq.~\ref{eq:LNAT}:
\begin{equation}
    \mathcal{L}_{NAT}^{*} = \mathcal{L}_{NAT}  +\lambda_2 \mathcal{L}_{IL}
\end{equation}
where $\lambda_1$ and $\lambda_2$ are hyper-parameters, which are set to 0.001 in our experiments.

%\paragraph{SPD for latent space layer}\quad We regard $q$ as an ideal latent space signal, and for the same reasons as above, it is hard to accurately generate the same sequence as $q$. We propose SPD(space parallel decoding for latent space layer) to obtain high-quality $u_i$. Different from the LPD for target length, we operate local sample that do not run through the decoder. Consider that the shape of $u$ is $T'\times|S|$, we sample $k_{sample}$ elements from latent space to get a new output, the shape of which is $T'\times k_{sample}$. At last we re-score the $k_{sample}$ latent sequences and select the best one. In this way, we limit the expansion of the computational process to the local part of the network, and avoid consuming too many computing resources. 

\section{Experiments}

We evaluate our proposed model on machine translation tasks and provide the analysis. We present the experimental details in the following, including the introduction to the datasets as well as our experimental settings. 

\paragraph{Datasets} 
We evaluate the proposed method on three widely used public machine translation corpora: IWSLT16 En-De(196K pairs), WMT14 En-De(4.5M pairs) and WMT16 En-Ro(610K pairs). All the datasets are tokenized by Moses~\citet{Koehn2007MosesOS} and segmented into $32k-$subword symbols with byte pair encoding~\citet{BPE} to restrict the size of the vocabulary. For WMT14 En-De, we use newstest-2013 and newstest-2014 as development and test set
respectively. For WMT16 En-Ro, we use newsdev-2016 and newstest-2016 as development and test sets respectively. For IWSLT16 En-De, we use test2013 as validation for ablation experiments.

\paragraph{Knowledge Distillation Datasets}
Sequence-level
knowledge distillation is 
applied to alleviate
multimodality in the training dataset, using
the AT demonstrator as the
teachers~\cite{kim_sequence-level_2016}. We replace the reference target sentence of each pair of training example $(X, Y)$ with a new target sentence $Y^*$, which is generated from the teacher model(AT demonstrator). Then we use the new dataset $(X, Y^*)$ to train our NAT model.
To avoid the redundancy of running fixed teacher models repeatedly on
the same data, we decode the entire training set once
using each teacher to create a new training dataset for its respective student.

\paragraph{Model Settings} We first train the AT demonstrator and then freeze its parameters during the training of \method. In order to speed up the convergence of NAT training, we also initialize \method\ with the corresponding parameters of the AT expert as they have similar architecture. 
For WMT14 En-De and WMT16 En-Ro, we use the hyperparameter settings of base Transformer model in \citet{transformer}($d_{model} = 512, d_{hidden} = 512, n_{layer} = 6 \quad and \quad n_{head} = 8$). As in \citet{gu_non-autoregressive_2017,lee_deterministic_2018}, we use the small model ($d_{model} = 278, d_{hidden} = 507, n_{layer} = 5\quad and\quad n_{head} = 2$) for IWSLT16 En-De. For sequence-level distillation, we set beam size to be 4.
For \method, we set the number of action category to $512$  and found \method is robust to the setting in our preliminary experiments.

\paragraph{Length Parallel Decoding}
For inference, we follow the common practice of noisy
parallel decoding ~\cite{gu_non-autoregressive_2017}, which generates a number
of decoding candidates in parallel and selects the best translation via re-scoring using AT teacher. In our scenario, we first train a module to predict the target length as $\hat{T}$. However, due to the inherent uncertainty of the data itself, it is hard to accurately predict the target length. A reasonable solution is to generate multiple translation candidates by predicting different target length $\in [\hat{T}-\Delta T,\hat{T}+\Delta T]$ , which we called LPD (length parallel decoding).  The model generates several outputs in parallel,  then we use the pre-trained autoregressive model to identify the best overall translation.

%%%%%%%%%%%%%%%%%%%%%%%%%%%%%%%%%
\begin{table*}[ht]
%\resizebox{\textwidth}{15mm}{
\centering
\begin{tabular}{c|ccccc|c}
\hline
&Distill    &UniformCopy       & SoftCopy                  & LPD        & Imitation Learning                & BLEU  \\
\hline  
1&           &$\surd$ &                          &            & w/o                           & 16.51 \\
2&$\surd$    &$\surd$ &                         &            & w/o                           & 20.72 \\%17.7%& 16    (13.5$\times$)  
3&$\surd$    &                  & $\surd$                  &            & w/o                           & 25.34 \\ %& 21.5  (16.3$\times$)
4&           &                  & $\surd$          &            & w/                   & 23.56 \\ %& 29.4  (12$\times$)  
%           &                  & $\surd$    &              &            & 3                   & 23.49 \\ %& 22.0  (16$\times$)  
5&           &                  & $\surd$          & $\surd$    &  w/                   & 24.35 \\ %& 76.7  (4.6$\times$) 
%6&$\surd$    &$\surd$           &                  &            & 3                   & 28.26 \\ %& 38.1  (9.2$\times$) 
%7&$\surd$    &$\surd$           &                          &            & 3                   & 28.27 \\ %& 26.8  (13)$\times$  
%$\surd$    &$\surd$           &            & $\surd$      & $\surd$    & 3    & $\surd$               & 29.18 \\ %& 96.4  (3.6$\times$) 
%$\surd$    &                  & $\surd$    &              & $\surd$    & 3    & $\surd$               & 30.73 \\ %& 44.7  (7.8$\times$) 
%$\surd$    &$\surd$           &            & $\surd$      & $\surd$    & 3    & $\surd$               & 30.34 \\ %& 78.9  (4.4$\times$) 
%$\surd$    &$\surd$           &            &              & $\surd$    & 3    & $\surd$               & 30.58 \\ %& 44.9  (7.8$\times$) 
%$\surd$    &                  & $\surd$    &              &            & 3                   & 28.41 \\ %& 21.3  (16.5$\times$)
6&$\surd$    &                  & $\surd$         &            &  w/                  & 28.41 \\ %& 31.6  (11$\times$)  
7&$\surd$    &                  & $\surd$          & $\surd$    &  w/                  & 30.68 \\ %& 76.7  (4.6$\times$) 
%$\surd$    &                  & $\surd$    &              &            & 3    &                       & 27.78 \\ %& 22.3  (15.7$\times$)
%$\surd$    &                  & $\surd$    &              &            & 3    & $\surd$               & 27.49 \\ %& 18.9  (18.6$\times$)
%$\surd$    &                  & $\surd$    & $\surd$      &            & 3    & $\surd$               & 27.87 \\ %& 29.6  (11.8$\times$)
\hline
\end{tabular}

\caption{Ablation study on the dev set of IWSLT16. \textit{w/} indicates \textit{with} and \textit{w/o} indicates \textit{without}. LPD indicates length parallel decoding. } \label{tab:iwslt}
\end{table*}
\section{Results and Analysis}

\paragraph{Competitor} We include three  NAT works as our competitors, the NAT with fertility (NAT-FT) \cite{gu_non-autoregressive_2017}, the NAT with iterative refinement (NAT-IR) \cite{lee_deterministic_2018} and  the NAT with discrete latent variables ~\cite{kaiser_fast_2018}. For all our tasks, we obtain the baseline performance by either directly using the performance figures reported in the previous works if they are available or producing them by using the open source implementation of baseline algorithms on our datasets. The results are shown in Table~\ref{tab:bleu}.

\paragraph{1. \method significantly improved the quality of the translation with a large margin.} 
On all the benchmark datasets, our \method with LPD achieves the best translation performance, which is even  close to the results of the autoregressive model, e.g. 30.68 vs. 30.85 on IWSLT16 En$\rightarrow$De tasks, and 31.81vs. 32.59 on WMT16 Ro$\rightarrow$En tasks. It is also worth mentioning that introducing the imitation module to AT demonstrator does not affect both the performance and the inference speed compared with the standard transformer model.

\paragraph{2. \method Imitation learning plays an important role on bridging the gap between \method and AT demonstrator} 
Clearly, \method  leads to remarkable improvements over the competitor without imitation module (over almost $3$ BLEU score on average). To make a fair comparison, the competitor 
follow exactly the same training steps with \method, including the initialization, knowledge distillation, and Soft-Copy.
The only difference comes from the imitation module.

\paragraph{3. \method gets  better latency.} For NAT-FT, a big sample size(10 and 100) is required to get satisfied results, which seriously affects the inference speed of the model. Both NAT-FT and NAT-IR, the efficiency of models with refinement technique drop dramatically($15.6\times\rightarrow2.36\times$ of NAT-FT and $8.9\times\rightarrow1.5\times$ of NAT-IR). Our  \method gets even better performance with faster speed. The speedup compared with AT model is $9.7\times$.

\subsection{Ablation Study}

To further study the effects brought by different techniques, we show in Table~\ref{tab:iwslt} the translation performance of different NAT model variants for the IWSLT16 En-De translation task.

%%%%%%%%%%%%%%%%%%%%%%%%%%%%%%%%%%
\paragraph{Soft-Copy v.s. Uniform-Copy}
The experimental results show that Soft-Copy is better than Uniform-Copy. Since Uniform-Copy employs a hard copy mechanism and   directly copies the source embeddings without considering  the global information, which increases the learning burden of the decoder. 
Our model takes the output of encoder as input and proposes a differentiable copy mechanism
which gets much better results(25.34 vs. 20.71, see in line 3 and 2). 

\paragraph{Imitation Learning v.s. Non Imitation Learning}
The imitation learning method  leads to an improvement of around 3 BLEU points(28.41 vs. 25.34, see line 6 and 3). NAT without IL degenerates into a normal NAT model. As discussed in section~\ref{sec:intro}, current NAT approaches suffer from delayed supervisions~(or rewards) and large search space in training. NAT decoder simultaneously generates all words of the translation, the search space of which is very large. 

\paragraph{Length Parallel Decoding}
Compared with the greedy beam search, LPD technique improves the performance around 2 BLEU points(30.68 vs. 28.41, from line 7 and 6). 
The observation is in consist with our intuition that 
sampling from the length space can improve the performance.

\paragraph{Complementary with Knowledge Distillation}
In consist with previous work, NAT models achieved +4.2 BLEU score from sequence level knowledge distillation technique (see in row 1 and row 2).  \method without knowledge distillation obtained 23.56 BLEU score  which is comparable to non-imitation NAT with knowledge distillation (see in row 3 and row 4).
More importantly, we found that the imitation learning framework complemented with knowledge distillation perfectly. 
As shown in row 3 and 6, \method substantially improves the performance of non-imitation NAT knowledge distillation up by +3.3 BLEU score.

%%%%%%%%%%%%%%%%%%%%%%%%%%%%%%%%
\begin{figure}[h]
\centering
\footnotesize
\includegraphics[width=1.0\linewidth]{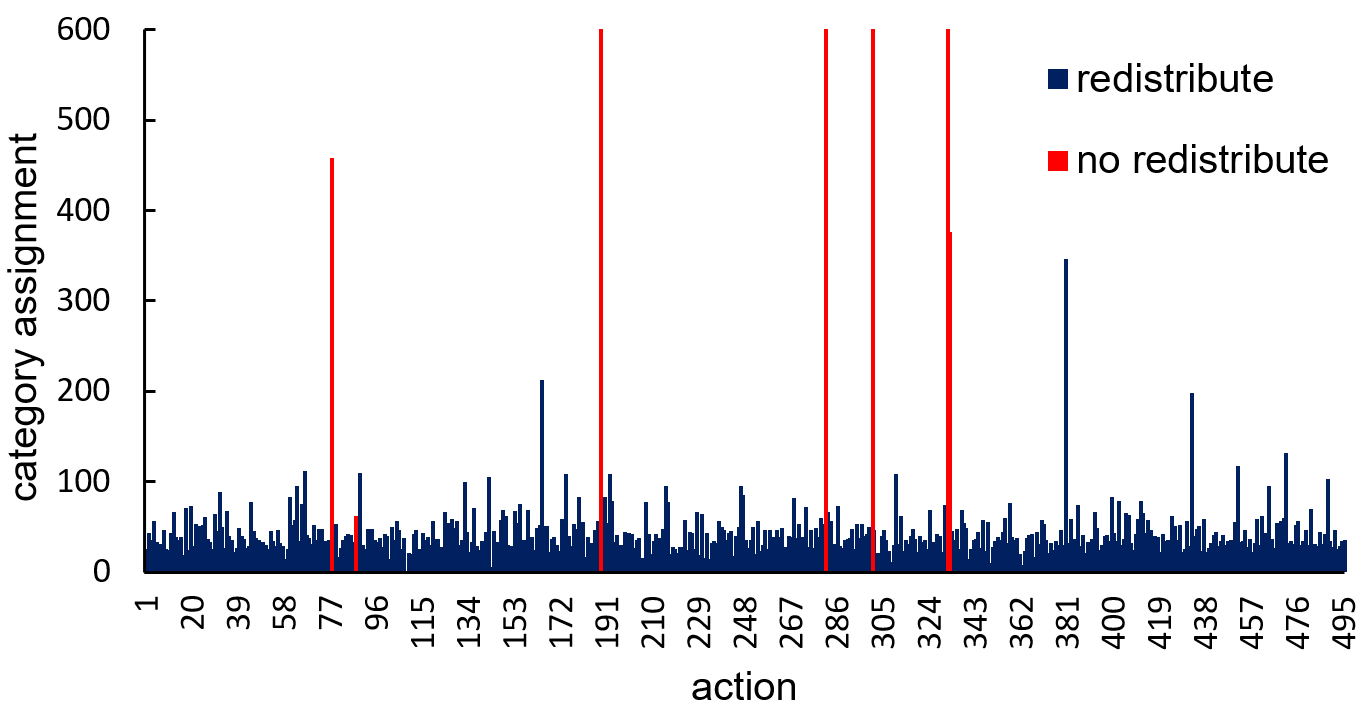} %re-crop.pdf
\caption{Action category assignment distribution. Redistribute method leads to a more balanced distribution(blue), otherwise, it will be extremely unbalanced(red). \label{fig:re}}

\end{figure}
%%%%%%%%%%%%%%%%%%%%%%%%%%%%%%%%%%%%%%%%%%%%%%
\paragraph{Action Distribution Study}
One common problem in unsupervised clustering is that the results are unbalanced. In this paper, we call that an action is selected or activated when its probability in $\pi(o_t)$ is maximum. Then  the space usage can be calculated by counting the number of times each action is selected. We evaluate the space usage on the development set of IWSLT16, and the results are presented in Figure~\ref{fig:re}. We greatly alleviate the problem of space usage through the category redistribution technique(Eq.\ref{eq:re1}, Eq.\ref{eq:re2}). When building the model without category redistribution, most of the space is not utilized, and the clustering results are concentrated in a few spatial locations, and the category information cannot be dynamically and flexibly characterized. In contrast, category redistribution makes the category distribution more balanced and more in line with the inherent rules of the language, so the clustering results can effectively guide the learning of the NAT model.

\section{Related Work}
 \citet{gu_non-autoregressive_2017} first developed a non-autoregressive NMT system which  produces the outputs in parallel and the inference speed is thus significantly boosted. However, it comes at the cost that the translation quality is largely sacrificed since the intrinsic dependency within the natural language sentence is abandoned. A bulk of work has been proposed to mitigate such performance degradation.
 \citet{lee_deterministic_2018} proposed a method of iterative refinement based on latent variable model
and denoising autoencoder. \citet{libovicky_end--end_2018} take NAT as a connectionist
temporal classification problem, which achieved better latency. 
~\citet{kaiser_fast_2018} use discrete latent variables that makes
decoding much more parallelizable.
They first auto encode the target sequence into a shorter sequence
of discrete latent variables, which at inference
time is generated autoregressively, and finally decode the output sequence from the shorter latent
sequence in parallel.
\citet{guo_nat_decinp}  enhanced decoder input by introducing phrase table in SMT and embedding transformation.
\citet{2019Auxiliary} leverage the dual nature of translation tasks (e.g., English to German and German to English) and minimize a backward reconstruction error to ensure that the hidden states of the NAT decoder are able to recover the source side sentence. 

Unlike the previous work to modify the NAT architecture or decoder inputs, we introduce an imitation learning framework to close the performance gap between NAT and AT. To the best of our knowledge,  it is the first time that imitation learning was applied to such problems.

\section{Conclusion}
We propose an imitation learning framework for non-autoregressive neural machine translation to bridge 
the performance gap between NAT and AT.
Specifically, We propose to employ a knowledgeable AT demonstrator to supervise every decoding state of NAT across different time steps and layers.
As a result, \method leads to remarkable improvements and largely closes the performance gap between NAT and AT on several benchmark datasets.

As a future work, we can try to improve the performance of the NMT by introducing more powerful demonstrator with different structure (e.g. right to left).  Another direction is to apply the proposed imitation learning framework to  similar scenarios such as simultaneous interpretation.

\section*{Acknowledgement}
We thank the anonymous reviewers for their thoughtful comments. Xu Sun is the corresponding author of this paper.

\bibliographystyle{acl_natbib}
\bibliography{acl2019}

\end{document}